\documentclass[position,final]{article}

\usepackage{neurips_2026}

\usepackage[utf8]{inputenc} 
\usepackage[T1]{fontenc}    
\usepackage{hyperref}       
\usepackage{url}            
\usepackage{booktabs}       
\usepackage{amsfonts}       
\usepackage{nicefrac}       
\usepackage{microtype}      
\usepackage{xcolor}         
\usepackage{amsmath}
\usepackage{amssymb}
\usepackage{mathtools}
\usepackage{amsthm}
\usepackage{wrapfig}

\let\citet\cite
\let\citep\cite

\theoremstyle{plain}

\theoremstyle{definition}

\theoremstyle{remark}

\usepackage[disable,textsize=tiny]{todonotes}

\newcommand{\E}[1]{\mathbb{E}\left[#1\right]}

\newcommand{\dd}{\text{\,d}}
\definecolor{Purpleee}{RGB}{91,40,224}

\DeclareMathOperator{\erfc}{erfc}
\DeclareMathOperator{\erf}{erf}

\title{Wasserstein Gradient Flows and Forward-Only \\ Diffusion Are Not Enough for Multimodal Sampling}

\author{
Daniel McBride, Pratik Khandagale, Cristina Garcia-Cardona, Yen Ting Lin\thanks{Corresponding author: yentingl@lanl.gov} \\\\ 
Los Alamos National Laboratory, Los Alamos, NM 87545, USA \\ 
}

\begin{document}

\maketitle

\begin{abstract}
There has been a proliferation of sampling algorithms based on Wasserstein gradient flows (WGF) and forward-only diffusion processes (FODP), often accompanied by theoretical guarantees of exponentially fast convergence to the target distribution. These guarantees are frequently interpreted as evidence that such methods can efficiently sample complex multimodal distributions, often supported by empirical results. In this work, we argue that this interpretation is fundamentally misleading. By invoking the Jordan--Kinderlehrer--Otto (JKO) scheme and Otto calculus, we establish that the canonical WGF sampling dynamics and overdamped forward diffusion share the same density evolution and therefore inherit the same metastability and slow-mixing phenomena long understood in nonequilibrium statistical physics. We analyze this family of samplers using two complementary tools---spectral analysis and mean first-passage time (MFPT) analysis---and show that well-separated multimodality can induce exponentially long mixing times associated with small spectral gaps and rare inter-mode transitions. For the commonly adopted log-linear annealing schedule studied here, we find that introducing intermediate distributions does not remove the exponential scaling of the total transport time. The limitation is structural rather than implementation-specific: purely local, gradient-driven transport mechanisms can require exponentially long times to transport probability mass across well-separated modes. We argue that this represents a fundamental limitation of WGF- and FODP-based sampling in their standard forms, and motivates future development of fundamentally nonlocal mechanisms for efficient multimodal sampling.
\end{abstract}

\section{Introduction and Background}\label{sec:intro}

We are interested in the \textit{sampling problem}: given a potentially unnormalized density $\tilde{\nu}(x)$ defined on a continuous state space $x \in \mathbb{R}^D$, the goal is to generate i.i.d.~samples from the corresponding normalized density function $\nu(x) = \tilde{\nu} (x)/ \mathcal{Z}$, where $\mathcal{Z} = \int \tilde{\nu}(x) dx$ is the (typically intractable) normalization constant. This problem is fundamental across a wide range of scientific domains, including statistical physics, molecular dynamics simulation, and Bayesian inference \cite{faulknerSamplingAlgorithmsStatistical2023,heninEnhancedSamplingMethods2022,wilkinsonBayesianMethodsBioinformatics2007,sharmaMarkovChainMonte2017,jackmanEstimationInferenceBayesian2000,mccartanSequentialMonteCarlo2023}.

A classical approach to this problem is Markov Chain Monte Carlo (MCMC; \citet{metropolisEquationStateCalculations1953}), which constructs a Markov chain whose stationary distribution matches the target distribution. While MCMC methods are asymptotically exact, their practical performance is often limited by slow mixing, particularly in high-dimensional ($D\gg 1$) or multimodal settings. Various extensions have been proposed to address the challenge, including Hamiltonian Monte Carlo \cite{duaneHybridMonteCarlo1987,nealMCMCUsingHamiltonian2012} and Sequential Monte Carlo \cite{del2006sequential}. A growing body of recent work has introduced alternative sampling methods which learn probability transport maps \cite{zhang2021path,vargas2023denoising,tian2024liouville}, as well as methods inspired by Wasserstein gradient flows \cite{liu2016stein,liu2017stein,wibisono2018sampling,berntonLangevinMonteCarlo2018,carrilloBlobMethodDiffusion2019,chenProjectedSteinVariational2020,mokrov2021large,korbaKernelSteinDiscrepancy2021,wang2022projected,ilin2025score,fanVariationalWassersteinGradient2022a} and diffusion processes \cite{robertsExponentialConvergenceLangevin1996,welling2011bayesian,girolamiRiemannManifoldLangevin2011,chenStochasticGradientHamiltonian2014,chengUnderdampedLangevinMCMC2018,durmusEfficientBayesianComputation2018}. We particularly focus on methods based on Wasserstein gradient flows (WGF) and forward-only diffusion processes. These approaches are often accompanied by theoretical guarantees of exponential convergence to the target distribution, which are sometimes interpreted as indicating fast practical convergence. We revisit this interpretation by studying the mixing behavior of such methods. Our central thesis is that, despite exponential convergence guarantees, practical performance can be severely limited in multimodal settings. In particular, when the target distribution exhibits well-separated modes, the dynamics---being inherently local and gradient-driven---lead to exponentially long mixing times, rendering efficient sampling computationally infeasible.

Our position is based on a unified analytical framework. We first show, using the Jordan--Kinderlehrer--Otto (JKO) scheme \cite{jordan1998variational} and Otto calculus \cite{otto2001geometry}, that canonical WGF sampling dyanmics and overdamped forward diffusion-based samplers share the same density evolution. This connection allows us to leverage classical analytical tools from nonequilibrium statistical physics, where the forward diffusion process has been thoroughly studied. We then apply two standard analytical tools developed therein, spectral analysis \cite{risken1989fokker,gardiner2004handbook,pavliotis2014stochastic} and mean first-passage time (MFPT) analysis \cite{van1992stochastic,hanggi1990reaction}, to characterize mixing behavior and demonstrate how small spectral gaps lead to practically slow convergence. Our work brings classical analytical tools from nonequilibrium statistical physics into the context of modern sampling methods, providing a complementary perspective on their practical limitations.

Our claim is not a universal impossibility result for multimodal sampling, nor do we claim that WGF- or forward diffusion-based methods fail for \emph{every} multimodal target. Instead, our critique concerns applying these methods, which rely solely on local, gradient-driven dynamics, to well-separated multimodal target densities. As shown below, in the metastable regime, these dynamics can require exponentially long times to transport probability mass between modes, leading to impractically slow convergence to the true target distributions.

We emphasize that the contribution of this position paper is not a new sampling method, benchmark, or theory of metastability. Rather, it is a critical synthesis that connects the marginal-density equivalence between WGF and forward diffusion with classical spectral and MFPT analyses, thereby making explicit that these samplers inherit the metastability and slow inter-mode transport long understood for diffusion processes. 

We next provide minimal background on these two families of methods, together with pointers to more comprehensive references. 

\subsection{Wasserstein Gradient Flow-Based Samplers} \label{sec:WGF-sampler}

We first introduce WGF-based samplers, referring to methods that explicitly build upon the theory of Wasserstein gradient flows, as well as related approaches such as Stein Variational Gradient Descent (SVGD, \citet{liu2016stein,liu2017stein,chenProjectedSteinVariational2020}), which exhibit similar mathematical structure without explicitly invoking the WGF formalism.

To give a short description, Wasserstein gradient flow is a dynamical model of the evolution of a probability density function, or representative samples thereof, along the steepest gradient direction in the space of probability measures equipped with the Wasserstein-$W_2$ metric, with respect to a cost functional \cite{ambrosio2005gradient,figalliInvitationOptimalTransport2023}. To be mathematically precise, let us consider a cost functional $\mathcal{F}: L^1 \left(\dd x \right) \rightarrow \mathbb{R}$, mapping the density function $\rho$ to a real value representing the ``energy'' of the distribution. Our goal is to identify the steepest descent direction of $\mathcal{F}$ in the space of normalized probability density functions. Applying a time derivative to the cost functional yields
\begin{equation}
    \frac{\text{d}\mathcal{F}\left\{\rho\left(\cdot,t\right)\right\}}{\text{d} t} = \int \frac{\delta \mathcal{F}\left\{\rho\left(\cdot,t\right)\right\}}{\delta \rho } (x,t) \frac{\partial \rho \left(x,t\right)}{\partial t} \text{d} x. \label{eq:dEdt}
\end{equation}
Next, assuming a flow field $v(x,t)$ induces this evolving distribution $\rho(x,t)$, substituting the generalized Liouville equation \citep{gerlich1973verallgemeinerte} (the ``continuity equation'' in the optimal transport literature) 
\begin{equation}
    \partial_t \rho \left(x,t\right)= -\nabla_x \cdot \left[v(x,t) \rho(x,t) \right],
\end{equation}
into Eq.~\eqref{eq:dEdt} and performing an integration by parts results in
\begin{equation}
   \frac{\text{d} \mathcal{F}}{\text{d} t} = \int v^T (x,t) \cdot \nabla_x \frac{\delta \mathcal{F}\left\{\rho\left(\cdot,t\right)\right\}}{\delta \rho (x,t)}  \rho(x,t) \, \text{d} x, \label{eq:dEdt2} 
\end{equation}
assuming vanishing boundary terms. As the right-hand side of the equation can be interpreted as an inner product of a vector function $v(\cdot,t)$ with the gradient of the functional derivative with respect to the induced measure $\rho\left(\cdot, t\right)$, the steepest descent direction is 
\begin{equation}
   v_\text{WGF} (x,t) = -\nabla_x \frac{\delta \mathcal{F}\left\{\rho\left(\cdot,t\right)\right\}}{\delta \rho (x,t)},
\end{equation}
which we will refer to as the WGF velocity field.

A popular choice of the cost functional is the Kullback--Leibler (KL) divergence of the evolving distribution from the target distribution $\nu$, i.e., $\mathcal{F}\{\rho\} := \text{KL}\left(\rho \Vert \nu\right)$. It is straightforward (see Appendix~\ref{app:KL-Renyi}) to perform the variational calculus $\delta \mathcal{F}/\delta \rho$ in the space of normalized density functions (i.e., subject to the constraint $\int \rho(x,t) \, \text{d}x = 1$) to obtain the WGF with respect to the KL divergence
\begin{equation}
    v_\text{WGF,KL} = \nabla_x \log \nu(x) - \nabla_x \log \rho(x,t). \label{eq:WGF_KL_v}
\end{equation}
We focus our exact marginal-equivalence analysis on the KL-induced WGF in Eq.~\eqref{eq:WGF_KL_v}, which is the canonical free-energy flow underlying many WGF-based sampling methods. Alternative WGF objectives, including the $\alpha$-divergence considered in Appendix~\ref{app:KL-Renyi}, generate related local, gradient-driven dynamics but do not necessarily yield the identical Fokker--Planck equation.

In WGF-based sampling, i.i.d.~samples are first drawn from an easy-to-sample initial distribution, such as an isotropic Gaussian. Then, the samples are transported through the Wasserstein gradient flow. How one specifically implements this transport diverges into two categories. One category of methods \cite{liu2016stein,liu2017stein,carrilloBlobMethodDiffusion2019,chenProjectedSteinVariational2020,ilin2025score} approximate the Wasserstein gradient flow at the particle level, for example by estimating the score function $\nabla_x \log \rho$ or some regularization thereof (e.g., via kernel regression or density smoothing). The estimation is then used to evolve the samples according to the WGF velocity field \textit{deterministically}: 
\begin{equation}
\dot{x}(t)=\nabla_x \log \nu(x) - \nabla_x \log \rho(x,t). \label{eq:WGF_flow_dynamics}
\end{equation} 
The second category adopts the optimization approach directly, leveraging the JKO scheme \cite{jordan1998variational,wibisono2018sampling,berntonLangevinMonteCarlo2018,mokrov2021large,fanVariationalWassersteinGradient2022a} and solving for \textit{optimal transport maps} in between prespecified discrete times. Then, the optimal transport maps are deployed to transport the i.i.d.~samples from the initial distribution to the target.  

As with other gradient flows, WGF specifies a descent direction but not a unique time parametrization; multiple dynamical systems can realize the same WGF path. For example, one could define $\dot{x}(t)=\beta\left(t\right) \left[\nabla_x \log \nu(x) - \nabla_x \log \rho(x,t)\right]$, where $\beta(t)$ is a time-dependent scaling factor. Also, WGF-based sampling relies on the evolving score $\nabla_x \log \rho$, which is often unavailable analytically and must be estimated from samples. This score estimation can be difficult in high dimensions. Transport-map approaches face a different challenge: they require architectural constraints, such as Input Convex Neural Networks (ICNNs), to learn gradients of convex potentials, but these constraints may limit expressivity and scalability in complex multimodal settings \cite{korbaKernelSteinDiscrepancy2021,mokrov2021large,fanVariationalWassersteinGradient2022a}.

\subsection{Forward-Only Diffusion Process-Based Samplers}

We now move on to introducing a class of samplers which rely on a forward-only diffusion process (FODP) resembling the It\^o stochastic differential equation (SDE)
\begin{equation}
  \text{d} X_t = \nabla_x \log \nu \left(X_t \right) \text{d}t + \sqrt{2}\, \text{d} W_t,\label{eq:fSDE} 
\end{equation}
where $W_t$ is the standard Wiener process. Notable examples of this class include Unadjusted Langevin Algorithm (ULA, \citet{robertsExponentialConvergenceLangevin1996}), Metropolis-Adjusted Langevin Algorithm (MALA, \citet{robertsExponentialConvergenceLangevin1996}), Riemannian Manifold Langevin Dynamics \cite{girolamiRiemannManifoldLangevin2011}, Stochastic Gradient Langevin Dynamics \cite{welling2011bayesian}, and related momentum-based \cite{chenStochasticGradientHamiltonian2014} and underdamped \cite{chengUnderdampedLangevinMCMC2018} dynamics, as well as proximal projection methods (Moreau-Yoshida ULA/MYULA, \citet{durmusEfficientBayesianComputation2018}). We refer to this class as \textit{forward-only} diffusion processes\footnote{We avoid using \textit{Langevin Dynamics} as it can represent a broader class of random processes \cite{vankampenLangevinlikeEquationColored1989}.} because sampling is performed by simulating a forward It\^o SDE with the target distribution as its invariant measure, in contrast to Generative Denoising Diffusion Models \cite{sohl-dicksteinDeepUnsupervisedLearning2015,hoDenoisingDiffusionProbabilistic2020,songScorebasedGenerativeModeling2021} which require both forward and reverse-time SDEs. 

The application of FODP is comparatively more straightforward than WGF-based samplers. I.i.d.~samples are drawn from an \textit{a priori} defined distribution and then evolved according to Eq.~\eqref{eq:fSDE} (more precisely speaking, samples of the initial distribution are transported along the generated sample paths of Eq.~\eqref{eq:fSDE}), potentially with additional Metropolis adjustments or proximal projections for ensuring the correct invariant distribution with finite-time stepping.

\subsection{Samplers Based on Other Mechanisms}

Developing statistical sampling methods is an ongoing area of research. For completeness, we list a selection of samplers based on other mechanisms to contrast WGF- and FODP-based approaches. 

We remark that this is a biased selection and by no means a comprehensive review of the field. Annealed Flow Transport Markov Chain Sampler (AFTMC, \citet{arbel2021annealed,matthewsContinualRepeatedAnnealed2022}) and Liouville Flow Importance Sampler (LFIS, \citet{tian2024liouville}) are based on the structure of Sequential Monte Carlo (SMC, \citet{del2006sequential}) and Annealed Importance Sampling (AIS, \citet{nealAnnealedImportanceSampling2001}), leveraging both annealing and importance sampling techniques. Another line of work formulates sampling as variational inference in the space of stochastic processes, connecting to optimal control (Path Integral Sampler (PIS) by \citet{zhang2021path}) and Schr\"odinger bridge formulations (Denoising Diffusion Sampler (DDS) by \citet{vargas2023denoising}). \citet{richterImprovedSamplingLearned2023a} and \citet{bridgeMatchingSampler2026} adopt this perspective, but differ in how the stochastic dynamics are parameterized and optimized, ranging from structured Schr\"odinger bridge constructions to directly learned drift functions. 

\section{ Theoretical Analysis } \label{sec:theory}

\subsection{ Consolidating WGF and FODP }

Although the canonical WGF considered here and FODP are often presented as fundamentally different frameworks, they are equivalent at the level of marginal distributions. This connection can be traced back to the Jordan--Kinderlehrer--Otto (JKO) scheme \cite{jordan1998variational, otto2001geometry}, which shows that the evolution of probability densities under WGF coincides with that governed by the Fokker--Planck equation (FPE) \cite{risken1989fokker}.

More precisely, while the underlying sample dynamics differ significantly---WGF (Eq.~\eqref{eq:WGF_flow_dynamics}) defines a deterministic flow, whereas FODP (Eq.~\eqref{eq:fSDE}) corresponds to a stochastic process---their induced marginal density $\rho(x,t)$ evolves according to the same FPE. This can be seen by deriving the generalized Liouville equation \cite{gerlich1973verallgemeinerte} for the deterministic WGF dynamics in Eq.~\eqref{eq:WGF_flow_dynamics}:
\begin{align}
    \partial_t \rho_\text{WGF} 
    ={}& -\nabla_x \cdot \left[ \left(\nabla_x \log \nu(x) - \nabla_x \log\rho_\text{WGF} \right) \rho_\text{WGF} \right] \nonumber \\
    ={}& -\nabla_x \cdot \left[ \nabla_x \log \nu(x)\,\rho_\text{WGF} \right] + \nabla_x^2 \rho_\text{WGF}, 
    \label{eq:FPE1}
\end{align}
under the regularity and positivity assumptions ensuring $\rho_\text{WGF} > 0$ so that $\log \rho_\text{WGF}$ is well-defined. 

On the other hand, the marginal density $\rho_\text{FODP}$ induced by FODP satisfies the classical FPE \cite{risken1989fokker}:
\begin{equation}
    \partial_t \rho_\text{FODP} 
    = -\nabla_x \cdot \left[ \nabla_x \log \nu(x)\,\rho_\text{FODP} \right] + \nabla_x^2 \rho_\text{FODP}, 
    \label{eq:FPE2}
\end{equation}
again assuming strict positivity. Therefore, under standard regularity conditions ensuring uniqueness of solutions to the FPE, if $\rho_\text{WGF}(x,0) = \rho_\text{FODP}(x,0)$, $\rho_\text{WGF}(x,t) = \rho_\text{FODP}(x,t)$, $\forall t \ge 0$. We emphasize that this equivalence holds only at the level of marginal distributions; the joint laws (path measures) induced by the processes in Eqs.~\eqref{eq:WGF_flow_dynamics} and \eqref{eq:fSDE} are, in general, different.

Historically, both formulations originate from this shared foundation: the JKO scheme was introduced as a variational time-discretization method for solving the FPE, from which the geometric interpretation of WGF later emerged. In more recent expositions, such as Sec.~\ref{sec:WGF-sampler}, the perspective is often reversed---starting from an optimization objective and deriving the flow---while the connection to the underlying FPE is less emphasized.

This marginal equivalence has been well understood in the literature since \cite{jordan1998variational}, while \citet{wibisono2018sampling} and \citet{carrilloBlobMethodDiffusion2019} also noted and leveraged the equivalence. Our contribution is to leverage this equivalence operationally: results on the evolution of marginal distributions for diffusion processes---extensively studied in nonequilibrium statistical physics---can be directly transferred to WGF. In particular, since convergence is typically characterized by functionals of the marginal distribution, such as $\mathrm{KL}(\rho(\cdot,t)\Vert \nu)$, it suffices to analyze the convergence behavior of the FODP in Eq.~\eqref{eq:fSDE}.

Throughout the remainder of this analysis, we work in an idealized setting and focus on the diffusion process Eq.~\eqref{eq:fSDE}. Our results can be framed as the ``best-case scenario'' in which any invoked errors and biases are assumed negligible. That is, we ignore errors and biases arising from finite-particle approximations, score estimation, numerical time discretization, approximate transport maps, kernel density estimation, Metropolis corrections, or other implementation-specific details. Our focus is solely on the intrinsic mixing behavior of the limiting marginal dynamics, which, by the equivalence established above, can be studied through the diffusion process \eqref{eq:fSDE}.

In the next two sections, we analyze the convergence behavior of the FODP using two complementary approaches: spectral analysis and mean first-passage time analysis. Our goal is to derive analytic expressions of the mixing dynamics converging to the target $\nu$ without simulating the processes.

\subsection{ Spectral Analysis and Eigenfunction Expansion } \label{sec:spectral}

The generator of the FODP \eqref{eq:fSDE} is $\mathcal{L}:= \nabla_x \log \nu(x) \cdot \nabla_x + \nabla_x^2 $, whose adjoint $\mathcal{L}^\dagger$ is the Fokker--Planck operator. Because $\mathcal{L}^\dagger$ admits a suitable realization in a weighted $L^2$ space\footnote{More precisely, in a weighted $L^2$ space associated with the invariant measure $\nu$, e.g., $L^2(\nu^{-1})$. The Fokker--Planck operator $\mathcal{L}^\dagger$ formally acts on $L^1(\mathrm{d}x)$, but we work in $L^2$ to enable spectral analysis. For general $\nu$, $\mathcal{L}$ is not self-adjoint in standard $L^2$, but can be made symmetric under an appropriate change of measure. Our numerical calculations do not invoke this transformation; instead, we directly discretize the Fokker--Planck operator and analyze the spectrum of the resulting discrete-state generator.}, we can apply standard spectral analysis \cite{SpectralAnalysisFokkerPlanck,gardiner2004handbook,pavliotis2014stochastic} to express the solution of the FPE in terms of its eigenvalues and eigenfunctions. More specifically, we are interested in expressing the solution of Eqs.~\eqref{eq:FPE1} and \eqref{eq:FPE2}, given the initial condition $\rho\left(x,0\right):=\rho_0(x)$, as the eigenfunction expansion
\begin{equation}
\rho(x,t) = \sum_{n} c_n e^{\lambda_n t} \psi_n(x), \label{eq:eigenExpansion}
\end{equation}
where $\psi_n(x)$ are the right eigenfunctions of $\mathcal{L}^\dagger$, $\lambda_n$ and $c_n$ are the associated eigenvalues and coefficients. We remark that the final solution can be expressed in the above form because $\mathcal{L}^\dagger$ admits a discrete spectrum \cite{villaniHypocoercivity2009,pavliotis2014stochastic} if the potential $V:=- \log \nu$ is strongly confining, meaning that $-\log \nu(x) \to \infty$ as $\Vert x \Vert \rightarrow \infty$ super-linearly, which holds for a broad class of confining target densities. Ordered by their real parts, the eigenvalues are known  to satisfy $0=\Re(\lambda_0) > \Re(\lambda_1) \ge \Re(\lambda_2) \ge \dots$ with $\lambda_0 = 0$, the corresponding eigenfunction $\psi_0(x)$ is proportional to the target density $\nu(x)$, and the convergence time can be estimated (up to constants) by the spectral gap, $T_\text{mix} \approx \vert \Re(\lambda_1)\vert^{-1}$. Already, we can see that as the system relaxes exponentially to $\psi_0$ under standard irreducibility and spectral gap conditions, the convergence could still be long if $\vert \Re(\lambda_1) \vert $ is small.

There exists no closed-form solution for $\lambda_1$ for general target density $\nu$. For our one-dimensional test problems, we instead rely on numerical methods to estimate $\lambda_1$, using eigenexpansion \eqref{eq:eigenExpansion}, which is more advantageous than directly simulating the FODP \eqref{eq:fSDE} or time-integrating the FPE numerically. The eigenfunction expansion analysis isolates the slowest decaying mode without requiring full simulation of the dynamics, enabling the study of timescales that are difficult to access via direct simulation. We leave more detailed description of the numerical analysis in Sec.~\ref{sec:numerical}. 

This spectral characterization is not restricted to one-dimensional systems. For higher-dimensional confining landscapes, the semiclassical analysis of Fokker--Planck operators and the associated Witten Laplacians provides a well-established framework for characterizing the small eigenvalues associated with metastable transitions \cite{nier2005hypoelliptic}. In the metastable regime, these results underpin the Eyring--Kramers law, according to which the inverse spectral gap---and equivalently the dominant transition timescale---grows exponentially with the relevant energy or quasipotential barrier, while the local geometry near the modes and intervening saddle points enters through the prefactor. Thus, the small-spectral-gap mechanism studied here is not intrinsically one-dimensional, although our one-dimensional Gaussian-mixture examples provide a particularly transparent setting in which it can be calculated and verified quantitatively.

The generality of the spectral gap characterization should not be confused with the scalability of its brute-force numerical computation. A grid-based discretization using $N$ points along each coordinate produces an operator with $\mathcal{O}\left(N^D\right)$ degrees of freedom, making direct estimation of the spectral gap impractical beyond very low dimensions, even when sparse eigensolvers are used. We use numerical eigendecomposition as a quantitatively precise tool for illustrating the small-spectral-gap mechanism in our one-dimensional test problems, rather than as a scalable diagnostic for general high-dimensional targets.

\subsection{ Mean First-Passage Time Analysis } \label{sec:MFPT}

Mean first-passage time (MFPT) analysis is a complementary way to study convergence time. To define MFPT, let us consider the state space of the FODP \eqref{eq:fSDE} as $\Omega$ and a target set $S \subset \Omega$. The MFPT, denoted by $ T_S \left(x\right)$, is the expected time for the process starting at $x\in \Omega$ to reach the set $S$ for the first time. More formally, denote the random hitting time by
\begin{equation}
    \tau_S := \inf \left \{ t\ge 0 \vert X_t \in S, X_0 = x \right \}.
\end{equation}
Then the MFPT is $T_S(x)=\mathbb{E} \left[\tau_S \right]$, which satisfies the backward Kolmogorov equation
\begin{equation}
    \mathcal{L} T_S(x) = -1, \quad x \in \Omega \setminus S, \label{eq:mfpt_eq}
\end{equation}
with the boundary condition $T_S(x) = 0$ for $x \in S$ \cite{van1992stochastic,gardiner2004handbook}.

MFPT provides a powerful way for understanding the timescales of stochastic processes, especially for metastable systems. As we are interested in characterizing the mixing time for multimodal distributions, the MFPT from one mode to another can be used to estimate the transition rates between modes. We restrict attention to the simplest setting of bimodal test problems. Denote the MFPT from mode 1 to 2 (resp.~2 to 1) by $T_{1\rightarrow 2}$ (resp.~$T_{2\rightarrow 1}$). When the modes are well-separated, under the standard Markov State Modeling (MSM) framework \cite{schutte2013metastability,husicMarkovStateModels2018,perezLongTimeDynamicsParallel2016a}, the transition rates can be approximated by the inverse of the MFPTs, i.e., $k_{1\rightarrow 2} \approx T_{1\rightarrow 2}^{-1}$ and $k_{2\rightarrow 1} \approx T_{2\rightarrow 1}^{-1}$. Furthermore, the mixing time can be approximated by the inverse of the sum of the transition rates, i.e.,
\begin{equation}
    T_\text{mix} \approx \left(T_{1\rightarrow 2}^{-1} + T_{2\rightarrow 1}^{-1}\right)^{-1}. \label{eq:mixing_time_mfpt}
\end{equation}
See \citet{ngoImprovingEstimationKoopman2023} for a detailed discussion on estimating the system's relaxation rates using MFPTs.

The MFPTs can be derived in one dimension with a closed-form analytic expressions, Eq.~\eqref{eq:app_mfpt_onesided}, derived and presented in Appendix~\ref{app:MFPT}. We will compare the prediction to that from the eigenvalue-based estimation in Sec.~\ref{sec:numerical}.

\section{Numerical Experiments and Results} \label{sec:numerical}

The theoretical analyses in Sec.~\ref{sec:theory} suggest that, even in regimes where convergence is exponential, the associated convergence time can still be impractically long. In this section, we design simple test problems to demonstrate this potential pitfall for WGF- and FODP-based samplers. We emphasize that the numerical experiments are intentionally designed to be (1) simple, showing that these samplers can fail even in highly idealized settings, (2) representative of multimodal structures commonly encountered in real-world applications, and (3) easy to generate ground-truth samples for validation. In other words, the test problems are not intended to constitute a comprehensive benchmark suite.

\begin{wrapfigure}[20]{r}{0.55\textwidth}
    \centering
    \includegraphics[width=0.85\linewidth]{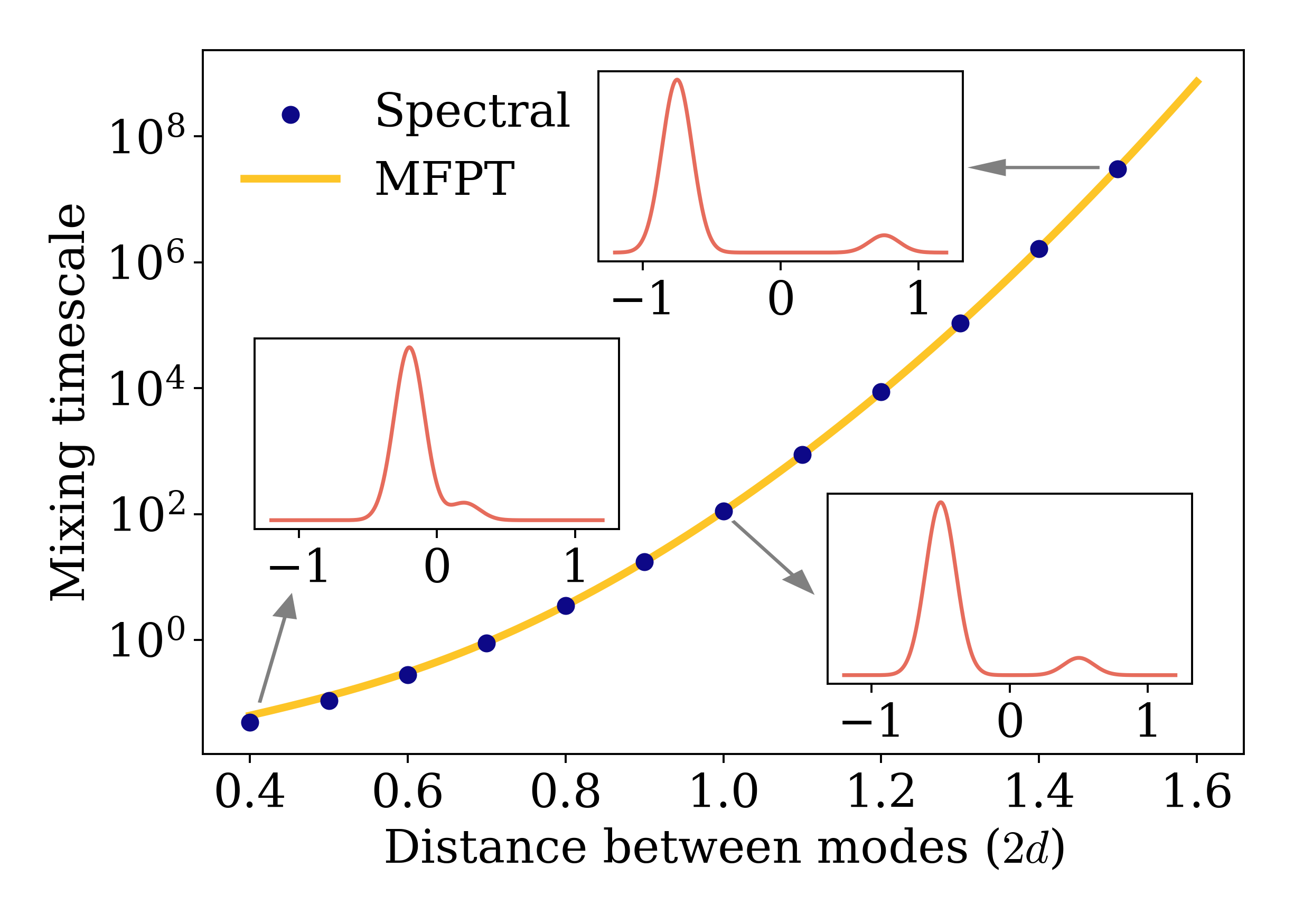}
    \caption{Mixing times predicted by the spectral (discrete markers) and MFPT (continuous line) analyses. Spectral: $\vert \Re \left(\lambda_1\right) \vert^{-1}$; MFPT: Eq.~\eqref{eq:mixing_time_mfpt} with Eq.~\eqref{eq:app_mfpt_gmm}. Insets illustrate the probability densities of the bimodal Gaussian mixtures.}
    \label{fig:mixingTime}
\end{wrapfigure}

The numerical analysis is structured as follows. We first specify the test problems, namely bimodal Gaussian mixtures. We begin with the one-dimensional case, where both eigenfunction expansion and MFPT analysis are computationally tractable, and compare their predictions. We then compare these theoretical estimates of the mixing time with direct simulations of either Eqs.~\eqref{eq:WGF_flow_dynamics} or \eqref{eq:fSDE}. Next, we demonstrate a sharp-transition behavior in several predictive metrics. The observed transition occurs precisely when the theoretically predicted timescale begins to exceed what is accessible via direct simulation. We will show that published methods also exhibit similar pathology. We further show that similar behavior persists in dimensions $D>1$, suggesting the same mechanism persists when embedded in a higher-dimensional state space. Finally, we examine whether the commonly used log-linear annealing schedule can resolve the issue and report negative results.

\subsection{Test problems} Throughout the rest of the section, we will focus on bimodal Gaussian mixtures:
\begin{equation}
    \nu(x) = w_1 \: \mathcal{N}(x; \mu_1, \sigma^2 I) + w_2 \: \mathcal{N}(x; \mu_2, \sigma^2 I). \label{eq:GT_distribution}
\end{equation}
The above expression is valid in a specified $D$-dimensional setting, where the centers $\mu_1$ and $\mu_2$ are $D$-dimensional vectors and $I$ is the $D\times D$ identity matrix. We fixed $\sigma^2=1.2\times 10^{-2}$, which sets the length scale of the Gaussian components. We will separate the modes along the first dimension by a distance $2d$, i.e., $\mu_1=\left[d, 0, \ldots \right]$ and $\mu_2 = \left[-d, 0, \ldots\right]$, where $d$ is a controlled parameter for us to study the mixing time. We will focus on asymmetric Gaussian mixture by fixing $w_1/w_2$ to be $1/10$ to showcase potential pitfall of the samplers.  

\subsection{Mixing time estimations in one dimension}

In $D=1$, both eigenfunction expansion and MFPT analysis are computationally tractable, allowing us to compare their predictions directly. For eigenfunction expansion, we first perform a uniform discretization in a truncated state space $[-1.2, 1.2]$ and utilize an upwind scheme \cite{lin2022koopman} for forming the discretized Fokker--Planck operator (FPO), documented in Appendix~\ref{app:discretization}. We perform eigen-decomposition of the discretized FPO and plot $\vert \Re\left(\lambda_1\right) \vert^{-1}$ in Fig.~\ref{fig:mixingTime}, as a function of the distance between the separated modes. On the same plot, we show the mixing timescale based on the MFPT analysis, i.e., Eq.~\eqref{eq:mixing_time_mfpt} with numerically integrated Eqs.~\eqref{eq:app_mfpt_gmm}. These two independent analyses show consistent results when the modes are well-separated ($2d \gtrsim 0.87$), which is the regime in which the two-state approximation underlying the MFPT analysis in Sec.~\ref{sec:MFPT} is valid.

The spectral and MFPT analyses show extremely stiff scaling as the mode separation increases; we remark that the $y$-scale is logarithmic. Even when the modes are not terribly separated (e.g., the right-most inset in Fig.~\ref{fig:mixingTime}), the mixing timescale exceeds $10^7$, where it is impractical to deploy the forward solver for sampling.

We next evaluate the FODP sampler by solving Eq.~\eqref{eq:fSDE} using the Euler--Maruyama algorithm, with initial conditions of the samples drawn from a standard Gaussian. We set the timestep as $10^{-3}$ and integrate until $T=10$. From Fig.~\ref{fig:mixingTime}, we estimate that the FODP should reasonably generate samples when $2d \lesssim 0.87$ but not when the modes are further separated. To quantify the performance of the sampler, we used the samples to evaluate (1) the approximate squared Wasserstein-2 distance between the FODP-generated samples and exact samples generated from the mixture directly, (2) the mean log probability of the terminal samples, and (3) the estimated left-mode weight, defined as the fraction of the samples in the left half-line ($x<0$). 

\begin{figure*}[!t]
    \centering
    \includegraphics[width=\linewidth]{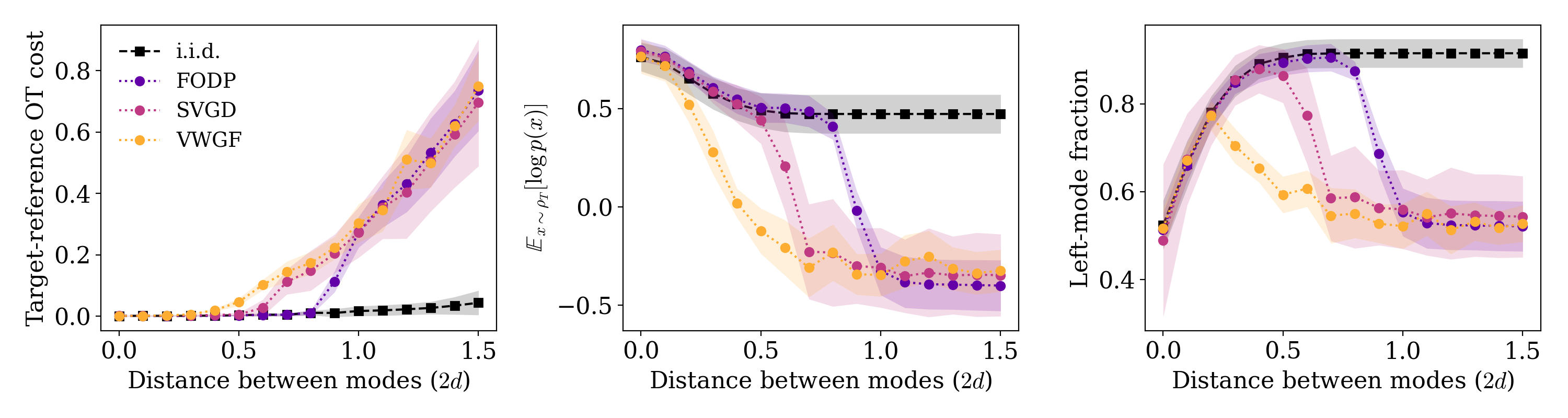}
    \caption{Performance comparison of different WGF-based samplers and i.i.d.~samples of the bimodal Gaussian mixture in $D=1$.} \label{fig:sampleQualities}
\end{figure*}

The results of the performance of the FODP sampler are presented in Fig.~\ref{fig:sampleQualities}, along with the same metrics of two other published WGF-based methods, SVGD and Variational Wasserstein Gradient Flow (VWGF, \citet{fanVariationalWassersteinGradient2022a}). Figure~\ref{fig:sampleQualitiesFODP} illustrates how this transition shifts as the terminal time $T$ increases for the FODP. In contrast, for SVGD and VWGF, increasing the terminal time has little effect on the metrics. These methods exhibit deterministic relaxation: once the particles have stabilized, more iterations do not meaningfully alter their configuration. We fixed the initial samples to be drawn from the Gaussian; more details of the experimental setup can be found in Appendix~\ref{app:experiment-specification}. We chose these methods because they are two notable representatives of the WGF-based sampling methods: SVGD evolves a kernelized approximation to the WGF velocity field in Eq.~\eqref{eq:WGF_flow_dynamics} and VWGF simulates WGF particle dynamics by invoking ICNN and the JKO scheme to solve the optimal transport problem. We observe that both the FODP- and WGF-based methods exhibit a ``phase transition\footnote{We are using this term as an analogy; this is not a phase transition in the strict statistical physics sense.}'' at sufficiently separated modes. We also observe that the transition point of FODP is consistent with the spectral and MFPT analyses ($2d \approx 0.87$). This transition corresponds to the point where the intrinsic relaxation timescale exceeds the accessible simulation horizon. This indicates that the quality of the generated samples sharply degrades beyond the predicted mixing timescale. Note that the tested particle methods degrade at smaller mode separations.

To illustrate that this feature is rather robust in higher dimensions, we present the same analysis for $D=10$ in Fig.~\ref{fig:sampleQualities10D}. In addition to similar qualitative behaviors, we remark that the critical transition mode separation distance predicted by the 1D analysis ($2d \approx 0.87$) is rather robust even for $D=10$. The agreement between one-dimensional MFPT predictions and higher-dimensional experiments can be understood through transition path theory \cite{eTowards2006,heymannGeometricMinimumAction2008,metznerTransitionPathTheory2009}. In metastable regimes, reactive trajectories concentrate near dominant minimum-action pathways passing through the relevant saddle points of the negative log-probability density landscape, so that the leading barrier-crossing mechanism can be described by an effectively one-dimensional reaction coordinate.

\subsection{A Log-Linear Annealing Schedule Does Not Alter the Observed Scaling} \label{sec:annealing}

\begin{wrapfigure}[20]{r}{0.55\textwidth}
    \centering
    \includegraphics[width=0.85\linewidth]{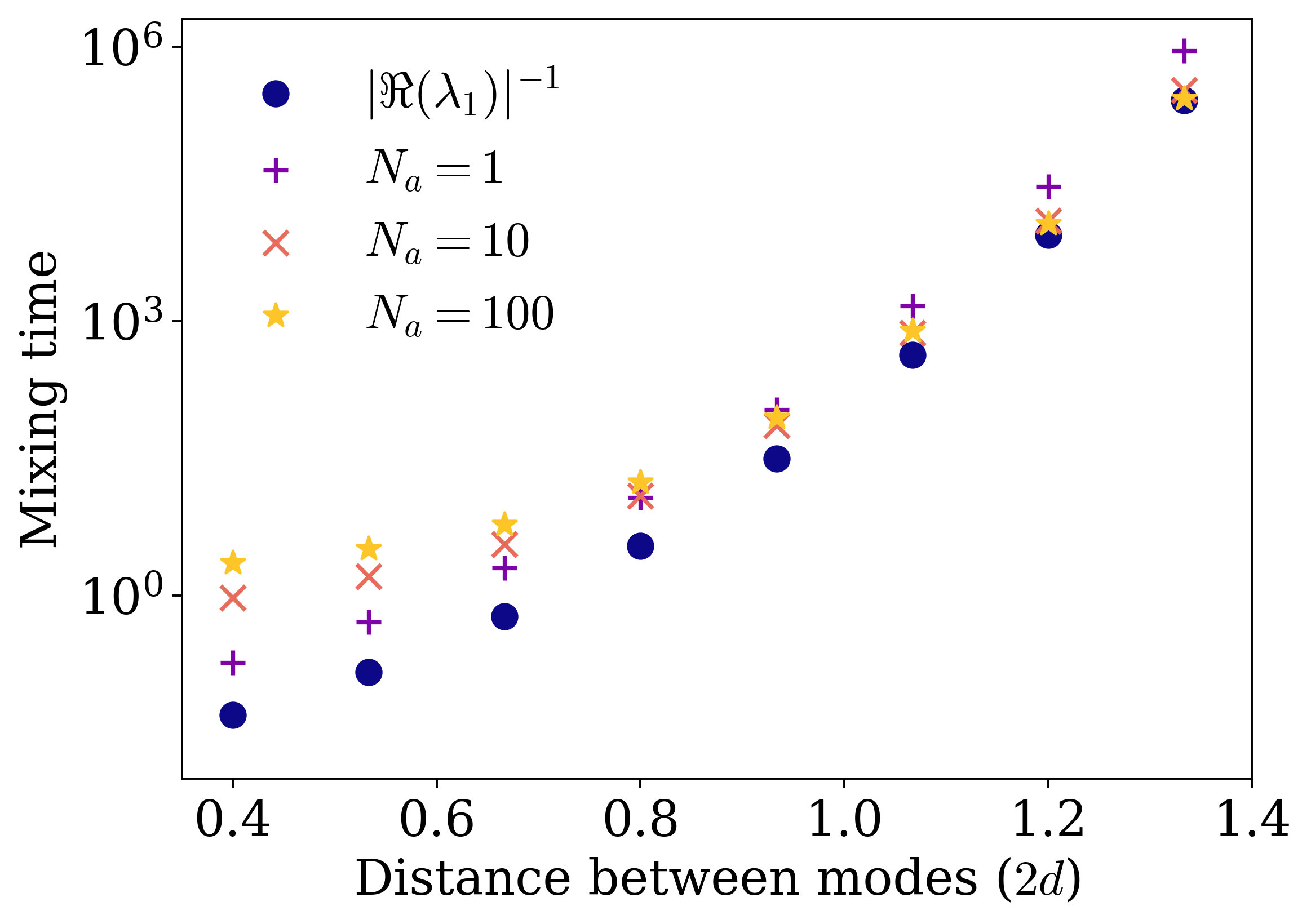}
    \caption{Mixing times when $N_a-1$ intermediate scales are used in annealing, measured by the description in Sec.~\ref{sec:annealing}. $N_a=1$ corresponds to no annealing. The complementary estimation using $\lambda_1$ presented in Sec.~\ref{sec:spectral} is shown for comparison, showing an accurate estimation of the scaling of mixing times for $2d\gtrsim 0.87$.}
    \label{fig:mixingTimeAnnealing}
\end{wrapfigure}
It has been suggested that annealing may improve the convergence of WGF-based samplers \cite{dangeloAnnealedSteinVariational2020,ilin2025score}. In this section, we apply spectral analysis and eigenfunction expansion to evaluate the performance of this approach on the idealized bimodal Gaussian mixture example in one dimension. As in Sec.~\ref{sec:theory}, we analyze the annealed WGF dynamics through the equivalent FODP.

We consider the typical annealing schedule which connects the initial density $\rho(\cdot, 0)$ and the target $\nu(\cdot)$ by $N_a$ transport stages, i.e., $\log \nu_k(x) = \left(1-{k}/{N_a}\right) \log \rho(x,0) + {k}/{N_a} \log \nu(x)$, where $k=0, 1, \ldots N_a$. Here, we analyze the typical geometric (log-linear) progression commonly used in annealing and importance sampling literature \cite{del2006sequential,tian2024liouville}. The annealed WGF-based sampling proceeds by sequentially transporting samples from scale $i$ to $i+1$ via WGF, until reaching the final scale. In between scales, it is assumed that WGF has evolved sufficiently long such that the samples approximate the next scale.

We ask how much \textit{total time} is needed for such a setup. As in Sec.~\ref{sec:theory}, we analyze the evolution of each intermediate distribution using the equivalent FODP and the eigenfunction expansion for the Fokker--Planck operator, i.e., Eq.~\eqref{eq:eigenExpansion}. Quantitatively, we define the sufficiency by the KL divergence $\text{KL}\left( \rho(t) \Vert \nu_{k+1} \right)\le \varepsilon$, where $\rho(0):=\nu_k$ and we fixed $\varepsilon=10^{-3}$. We adopt the forward KL divergence as a convergence criterion, as it is consistent with the KL-induced WGF formulation analyzed in Sec.~\ref{sec:WGF-sampler}. The required time can be computed numerically by expressing the forward solution in the form of the eigenexpansion \eqref{eq:eigenExpansion} without generating the actual sample paths. We compute the minimal time needed to satisfy the condition for each scale, and sum them up as an estimate of the ``mixing time'' needed for annealed WGF-based sampling. Note that even when $N_a=1$, the time estimated in this way is generally longer than the estimated timescale based on $\vert \Re(\lambda_1) \vert^{-1}$, which quantifies the longest ``half-life'' of the relaxing eigenmodes.

Our setup corresponds to an idealized best-case scenario in which each intermediate distribution is fully equilibrated before proceeding to the next scale. In practice, finite-time evolution and approximation errors would further increase the total computational cost.

Figure \ref{fig:mixingTimeAnnealing} shows the results from this analysis. 
We observe that while the log-linear annealing schedule can reduce the effective barrier between adjacent intermediate distributions, it does not fundamentally alter the exponential scaling of the overall mixing time with respect to mode separation. Instead, the total computational cost is redistributed across multiple intermediate stages. In particular, achieving successful transport requires sufficiently long evolution at each scale, leading to an overall cost comparable to that predicted by the spectral and MFPT analyses for the original target distribution. This suggests that the commonly adopted log-linear annealing schedule alone does not resolve the intrinsic slow-mixing behavior of local, gradient-driven dynamics. 

We conjecture that changing the intermediate path cannot systematically eliminate the metastable bottleneck when every stage continues to rely solely on the same local WGF/FODP dynamics. Establishing this conjecture across general annealing schedules, initial distributions, and target densities would require an optimality or lower-bound result beyond the scope of this work.

\section{Discussion and Outlook} \label{sec:discussion}

The central message of this work is that \textit{local, gradient-driven dynamics are not, by themselves, a generally efficient solution for sampling well-separated multimodal distributions}. Both WGF-based samplers and forward-only diffusion processes evolve probability mass using only local information, namely the score function $\nabla \log \nu(x)$. While such dynamics guarantee convergence in the asymptotic sense, our analysis shows that the associated mixing times can be exponentially long when the target distribution exhibits well-separated modes. This discovery is not new: it has long been recognized in the study of Langevin dynamics and diffusion processes, where transitions between metastable states are known to occur on exponentially long timescales governed by large-deviation rate functions and spectral gaps \cite{bovier2005metastability,hanggi1990reaction}. 

Our contribution is to connect this classical understanding to the analysis of Wasserstein-gradient-flow sampling. By establishing the equivalence between WGF and forward diffusion at the level of marginal distributions, we show that the idealized KL-induced WGF inherits the metastability and slow-mixing behavior already understood for overdamped diffusion. This connection allows spectral and MFPT analyses from statistical physics to be applied directly to its marginal convergence.

A key implication of our results is that \textbf{\textit{Wasserstein gradient flow, when used as a sampling mechanism in its current form, may represent a practical dead-end for general multimodal sampling}}. This statement should be interpreted with care. WGF remains a powerful theoretical framework for understanding optimization in the space of probability measures. However, when used to construct \textit{sampling algorithms}, it is based on \text{local information} to transport the samples and shares the same barriers associated with diffusion dynamics. Improvements in numerical discretization, particle approximation, parameterization (e.g., kernel methods or neural transport maps), or combining annealing, \textbf{do not fundamentally change} this behavior.

\begin{figure*}[!t]
    \centering
    \includegraphics[width=\linewidth]{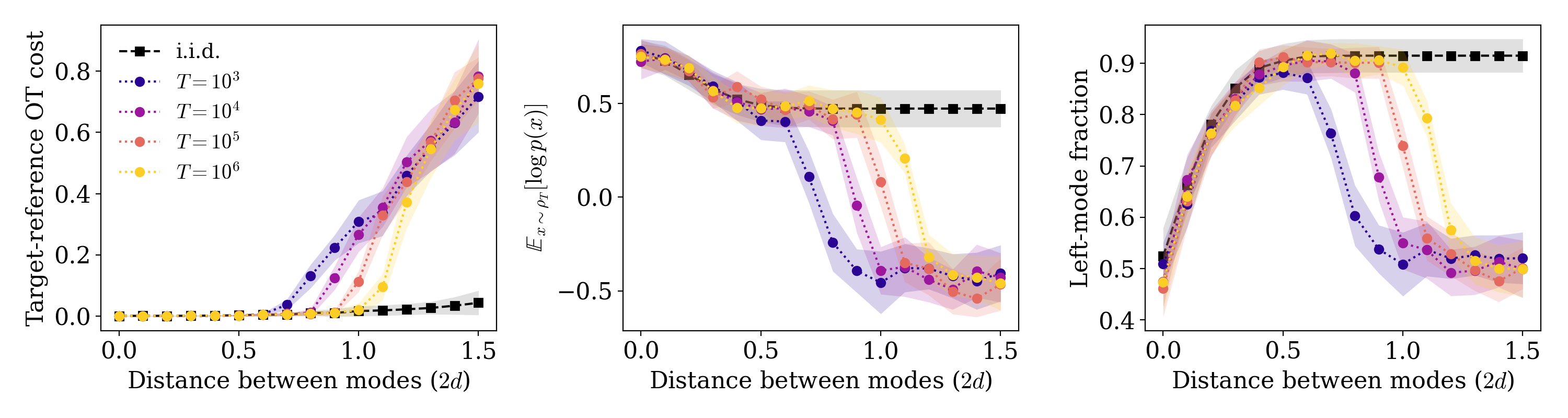}
    \caption{Performance comparison of FODP with different terminal time $T$ and i.i.d.~samples of the bimodal Gaussian mixture in $D=1$.}\label{fig:sampleQualitiesFODP}
\end{figure*}

Deterministically transported particle methods such as SVGD and VWGF exhibit an even sharper limitation than the stochastic FODP. For FODP, the intrinsic stochasticity provides a mechanism for rare but essential inter-mode transitions: over sufficiently long horizons, a small fraction of trajectories can escape local basins and cross low-probability regions. This is precisely the transport mechanism captured by the slowest decaying eigenmode ($\lambda_1$) of the Fokker--Planck operator and, in metastable regimes, is governed by large-deviation events. As illustrated in Fig.~\ref{fig:sampleQualitiesFODP}, increasing the terminal time $T$ can in principle improve the global statistics of FODP samples at a fixed mode separation, although the required time may be exponentially large. In contrast, deterministic WGF-based particle methods do not possess such a rare-event mechanism. Their evolution is purely ``relaxing'': particles are transported along deterministic velocity fields and eventually stabilize once the local flow field vanishes or becomes numerically negligible. After this stabilization, simply extending the time horizon does not create inter-mode crossings or improve the recovered mode weights. Thus, while FODP suffers from exponentially slow mixing, deterministic WGF-based samplers may fail more severely in practice: they can become trapped in locally equilibrated particle configurations that no longer explore the global structure of the target distribution. We emphasize that this observation concerns the tested finite-particle implementations. For the exact population-level KL-induced WGF in Eq.~\eqref{eq:WGF_flow_dynamics}, the marginal density follows the same Fokker--Planck equation as the FODP, and inter-mode mass transfer therefore occurs on the same slow timescale. 

Our analysis also sheds light on the role of annealing. It is believed that annealing can reduce the effective barriers between successive intermediate distributions. Our analysis shows that even under an idealized best-case assumption where each intermediate distribution is fully equilibrated, the total time required remains comparable to that predicted by the spectral and MFPT analyses for the original target distribution. This suggests that \textit{the commonly adopted log-linear annealing alone does not resolve the intrinsic slow-mixing behavior of local, gradient-driven dynamics}. In our experiment, the improvement obtained from introducing intermediate distributions is accompanied by additional transport time across the annealing stages, without changing the observed asymptotic scaling.

This observation naturally leads to a broader question: what distinguishes methods that succeed in sampling complex multimodal distributions from those that fail? Recent approaches such as AFTMC, PIS, DDS, and LFIS appear to perform well in regimes where WGF and FODP struggle. Despite their diverse formulations, these methods share a common feature: they incorporate \textit{nonlocal mechanisms} that go beyond local gradient transport. AFTMC combines learned transport maps with resampling, enabling particles to be relocated across distant regions of the state space. PIS and DDS operate in path space, optimizing over entire trajectories rather than applying local updates, effectively enabling coordinated global transport. In these formulations, the sampling problem is lifted to the path space, where the target distribution is enforced through global constraints or terminal conditions. LFIS, which relies on deterministic flows to transport samples, has a global coupling between all the samples through the logarithm of the partition function. These mechanisms provide ways to bypass the barriers that arise in purely local dynamics.

At a conceptual level, these observations suggest a unifying principle: \textit{efficient sampling of multimodal distributions requires mechanisms that utilize nonlocal information}. Whether achieved through resampling, importance weighting, or path-space optimization, the key departure from WGF and FODP lies in breaking the strict locality of gradient-driven updates. Understanding precisely how these mechanisms overcome metastability remains an open problem. For instance, while path-space formulations provide greater flexibility, their success depends on the choice of reference process and parameterization, and a systematic theory for designing such processes is still lacking.

Looking forward, two directions merit further investigation. First, it would be valuable to develop a more systematic framework for characterizing nonlocality in sampling algorithms and relating it to mixing complexity. Second, hybrid approaches that combine local gradient dynamics with controlled nonlocal moves may offer a promising compromise between stability and efficiency.

\newpage

\section*{ Author Contributions }
Y.T.L. and C.G.C.~conceived the project and designed the research; Y.T.L.~performed the theoretical analysis; all authors performed numerical analyses; Y.T.L.~wrote the paper and all authors reviewed, revised, and approved the manuscript.

\section*{ Acknowledgements }
This study was supported by the Laboratory Directed Research and Development (LDRD) program, under the project ``\textit{Renovating Monte Carlo Methods and Codebases with Generative Models}'' (20250391ER) at the Los Alamos National Laboratory (LANL). YTL was also partially supported by LDRD project ``\textit{Assured Artificial Intelligence}'' (20250048DR). This work is coauthored by employees of Triad National Security, LLC which operates LANL under Contract No.~89233218CNA000001 with the U.S.~Department of Energy and National Nuclear Security Administration. Y.T.L.~thanks Prof.~A.~Doucet for an insightful discussion during SIAM Conference on UQ, 2026.

\bibliographystyle{unsrt}

\appendix
\onecolumn

\section{Variational derivatives of KL and R\'enyi divergences}
\label{app:KL-Renyi}

In this appendix, we derive the Wasserstein gradient flow velocities corresponding to the Kullback--Leibler (KL) divergence and the R\'enyi $\alpha$-divergence. Throughout, we work in the space of normalized probability density functions, i.e.,
\begin{equation}
    \int \rho(x,t)\,\dd x = 1,
\end{equation}
so that admissible perturbations $\delta \rho$ satisfy
\begin{equation}
    \int \delta \rho(x)\,\dd x = 0.
\end{equation}

\subsection{KL divergence}
For the KL divergence,
\begin{equation}
    \mathcal{F}_{\mathrm{KL}}\{\rho\}
    :=
    \mathrm{KL}(\rho\Vert \nu)
    =
    \int \rho(x)\log\frac{\rho(x)}{\nu(x)}\,\dd x,
\end{equation}
consider the functional variation $\rho \mapsto \rho + \delta \rho$, $\delta \rho \ll 1$. Then
\begin{align}
    \mathcal{F}_{\mathrm{KL}}\{\rho + \delta \rho\} - \mathcal{F}_{\mathrm{KL}}\{\rho\}  &= \int \left(\rho + \delta \rho\right)\log\frac{\rho + \delta \rho}{\nu}\,\dd x - \int \rho \log\frac{\rho}{\nu}\,\dd x.
\end{align}
Collecting terms up to $\mathcal{O}(\delta \rho)$ 
\begin{align}
    \mathcal{F}_{\mathrm{KL}}\{\rho + \delta \rho\} - \mathcal{F}_{\mathrm{KL}}\{\rho\} ={}& \int \left(\log\frac{\rho(x)}{\nu(x)} + 1\right)\delta \rho(x)\,\dd x = \int \log\frac{\rho(x)}{\nu(x)} \delta \rho(x)\,\dd x, 
\end{align}
where the last equation was established by the normalization constraint $\int \delta \rho(x)\,\dd x = 0$. Thus, 
\begin{equation}
    \frac{\delta \mathcal{F}_{\mathrm{KL}}}{\delta \rho}(x) = \log \rho(x) - \log \nu(x).
\end{equation}
Therefore, the Wasserstein gradient flow velocity is
\begin{equation}
    v_{\mathrm{WGF,KL}}(x,t) = -\nabla_x \frac{\delta \mathcal{F}_{\mathrm{KL}}}{\delta \rho}(x) = \nabla_x \log \nu(x) - \nabla_x \log \rho(x,t).
\end{equation}

\subsection{R\'enyi $\alpha$-divergence}
For the R\'enyi $\alpha$-divergence,
\begin{equation}
    \mathcal{F}_{\alpha}\{\rho\}
    :=
    R_\alpha(\rho\Vert \nu)
    =
    \frac{1}{\alpha(1-\alpha)}
    \left(
    1 - \int \rho^\alpha(x)\nu^{1-\alpha}(x)\,\dd x
    \right),
    \qquad \alpha \in (0,1).
\end{equation}
Consider the functional variation $\rho \mapsto \rho + \delta \rho$, $\delta \rho \ll 1$. Then
\begin{align}
    \mathcal{F}_{\alpha}\{\rho + \delta \rho\} - \mathcal{F}_{\alpha}\{\rho\}
    ={}&
    \frac{1}{\alpha(1-\alpha)}
    \left(
    \int \rho^\alpha(x)\nu^{1-\alpha}(x)\,\dd x
    - \int (\rho(x)+\delta\rho(x))^\alpha\nu^{1-\alpha}(x)\,\dd x
    \right).
\end{align}
Collecting terms up to $\mathcal{O}(\delta \rho)$,
\begin{align}
    \mathcal{F}_{\alpha}\{\rho + \delta \rho\} - \mathcal{F}_{\alpha}\{\rho\}
    ={}&
    -\frac{1}{1-\alpha}
    \int
    \rho^{\alpha-1}(x)\nu^{1-\alpha}(x)\delta\rho(x)\,\dd x.
\end{align}
Thus,
\begin{equation}
    \frac{\delta \mathcal{F}_{\alpha}}{\delta \rho}(x)
    =
    -\frac{1}{1-\alpha}
    \rho^{\alpha-1}(x)\nu^{1-\alpha}(x).
\end{equation}
Therefore, the Wasserstein gradient flow velocity is
\begin{align}
    v_{\mathrm{WGF},\alpha}(x,t) ={}& \rho^{\alpha-2}(x,t)\nu^{-\alpha}(x)\left[\rho (x,t)\,\nabla_x \nu(x) - \nu(x)\,\nabla_x \rho(x,t)\right] \nonumber \\ ={}& \left( \frac{\rho(x,t)}{\nu(x)}\right)^{\alpha-1} \left[\nabla_x \log \nu(x) - \nabla_x \log \rho(x,t)\right].
\end{align}

\subsection{Remarks}
First, in the limit $\alpha \uparrow 1$, the R\'enyi divergence recovers the KL divergence, and correspondingly
\begin{equation}
    \lim_{\alpha \uparrow 1}
    v_{\mathrm{WGF},\alpha}(x,t)
    =
    \nabla_x \log \nu(x) - \nabla_x \log \rho(x,t)
    =
    v_{\mathrm{WGF,KL}}(x,t).
\end{equation}

Second, if the target density is specified only up to normalization, i.e.\ $\nu(x)=\tilde{\nu}(x)/\mathcal{Z}$, then
\begin{equation}
    \nabla_x \log \nu(x)
    =
    \nabla_x \log \tilde{\nu}(x),
\end{equation}
since the normalization constant $\mathcal{Z}$ is independent of $x$. Therefore, the Wasserstein gradient flow velocity can be evaluated using only the score of the unnormalized target density.

\section{Mean First-Passage Time in One Dimension} \label{app:MFPT}

In this appendix, we derive the one-dimensional MFPT expression used in Sec.~\ref{sec:MFPT}. Consider the one-dimensional FODP
\begin{equation}
    \dd X_t = \frac{\dd}{\dd x} \log \nu(X_t)\,\dd t + \sqrt{2}\,\dd W_t,
\end{equation}
with backward generator
\begin{equation}
    \mathcal{L} = \frac{\dd^2}{\dd x^2} + \frac{\dd}{\dd x} \log \nu(x)\,\frac{\dd}{\dd x}.
\end{equation}
Let the state space be $\Omega=\mathbb{R}$ and define the target set
\begin{equation}
    S := \{x \in \mathbb{R} \mid x \le a \text{ or } x \ge b\}, \quad a<b,
\end{equation}
so that the interior domain is $A=(a,b)=\Omega\setminus S$. The first hitting time is
\begin{equation}
    \tau_S(x) := \inf \{ t \ge 0 \mid X_t \in S, \; X_0 = x \},
\end{equation}
and the MFPT is $T_S(x)=\E{\tau_S(x)}$. It is well known that $T_S(x)$ satisfies the backward Kolmogorov equation
\begin{equation}
    \mathcal{L} T_S(x) = -1, \quad x \in (a,b), \quad T_S(a)=T_S(b)=0.
\end{equation}

Using $\frac{\dd}{\dd x} \log \nu(x)=\frac{\nu'(x)}{\nu(x)}$, the equation becomes
\begin{equation}
    -1 = \frac{\dd^2 T_S(x)}{\dd x^2} + \frac{\nu'(x)}{\nu(x)} \frac{\dd T_S(x)}{\dd x} = \frac{1}{\nu(x)} \frac{\dd}{\dd x} \left[ \nu(x) \frac{\dd T_S(x)}{\dd x} \right].
\end{equation}
Multiplying by $\nu(x)$ and integrating once gives
\begin{equation}
    C_1 - \int_{-\infty}^x \nu(x')\,\dd x' = \nu(x) \frac{\dd T_S(x)}{\dd x}.
    \label{eq:app_first_integral}
\end{equation}
Integrating again gives
\begin{equation}
    T_S(x) = C_2 + C_1 \int_a^x \frac{1}{\nu(x')}\,\dd x' - \int_a^x \frac{1}{\nu(x')}\,\dd x' \int_{-\infty}^{x'} \nu(x'')\,\dd x''.
    \label{eq:app_general_mfpt}
\end{equation}
The boundary condition $T_S(a)=0$ gives
\begin{equation}
    C_2 = 0.
\end{equation}
The boundary condition $T_S(b)=0$ then gives
\begin{equation}
    C_1 = \frac{ \int_a^b \frac{1}{\nu(x')}\,\dd x' \int_{-\infty}^{x'} \nu(x'')\,\dd x'' }{ \int_a^b \frac{1}{\nu(x')}\,\dd x' }.
    \label{eq:app_C1_absorb}
\end{equation}
Equations~\eqref{eq:app_general_mfpt}--\eqref{eq:app_C1_absorb} provide the MFPT for reaching the target set $S=\mathbb{R}\setminus(a,b)$.

For mode-to-mode transitions, we consider a one-sided limiting form. Suppose we seek the MFPT from a right mode located at $x$ to a left mode located at $a$ ($a<x$). We send the upper boundary $b \rightarrow \infty$. It can be proved following the analysis in \citep{linGeneExpressionDynamics2016} that $\lim_{b\rightarrow \infty} C_1 =1$. Equivalently, one may consider an alternative derivation by imposing a \textit{reflecting boundary condition} at $b$ before taking the limit. Under the reflecting condition $T_S'(b)=0$, Eq.~\eqref{eq:app_first_integral} gives
\begin{equation}
    C_1 = \int_{-\infty}^{b} \nu(x')\,\dd x' = 1.
\end{equation}
In either derivation, the limiting set $S=\left\{x<a\right\}$ and limiting value $C_1=1$ yields
\begin{equation}
    T_{\left\{x<a\right\}}(x) = \int_a^x \frac{1}{\nu(x')} \left[1-\int_{-\infty}^{x'} \nu(x'')\,\dd x'' \right]\dd x'.
\end{equation}
Defining the cumulative distribution function
\begin{equation}
    F(x) := \int_{-\infty}^{x} \nu(x')\,\dd x',
\end{equation}
we obtain
\begin{equation}
    T_{\left\{x<a\right\}}(x) = \int_a^x \frac{1-F(x')}{\nu(x')}\,\dd x'. \label{eq:app_mfpt_onesided}
\end{equation}
The left-to-right MFPT can be obtained analogously by reversing the coordinate direction.

For the two-component Gaussian mixture where the first mode is at $x=d$ and the second one at $x=-d$, 
\begin{equation}
    \nu(x) = w_1 \mathcal{N}(x;d,\sigma^2) + w_2 \mathcal{N}(x;-d,\sigma^2), \quad w_1+w_2=1, \label{eq:app_gmm}
\end{equation}
the cumulative distribution function is
\begin{equation}
    F(x) = \frac{w_1}{2} \left[ 1+\erf\left(\frac{x-d}{\sqrt{2}\sigma}\right) \right] + \frac{w_2}{2} \left[ 1+\erf\left(\frac{x+d}{\sqrt{2}\sigma}\right) \right].
\end{equation}
Thus,
\begin{equation}
    1-F(x) = \frac{w_1}{2}\erfc\left(\frac{x-d}{\sqrt{2}\sigma}\right) + \frac{w_2}{2}\erfc\left(\frac{x+d}{\sqrt{2}\sigma}\right),
\end{equation}
and Eq.~\eqref{eq:app_mfpt_onesided} becomes
\begin{subequations} \label{eq:app_mfpt_gmm}
\begin{align}
    T_{1 \to 2} ={}& \int_{-d}^d \frac{
    w_1 \erfc\left(\frac{x'-d}{\sqrt{2}\sigma}\right)
    +
    w_2 \erfc\left(\frac{x'+d}{\sqrt{2}\sigma}\right)
    }{2\nu(x')}\,\dd x',\\
    T_{2 \to 1} ={}& \int_{-d}^{d} \frac{1}{\nu(x')} \left[
    1-\frac{w_1}{2}\erfc\left(\frac{x'-d}{\sqrt{2}\sigma}\right)
    -\frac{w_2}{2}\erfc\left(\frac{x'+d}{\sqrt{2}\sigma}\right)
    \right]\dd x'. 
\end{align}
\end{subequations}
Here, consistent with Eq.~\eqref{eq:app_gmm}, label $1$ denotes the right mode and label $2$ denotes the left mode.

In practice, Eqs.~\eqref{eq:app_mfpt_gmm} can be evaluated using numerical quadrature. For well-separated modes, $d\gg\sigma$, the integral is dominated by the low-density barrier region between the modes, where $\nu(x')$ is exponentially small. As a result, the MFPT grows exponentially with the barrier height, providing the MFPT analogue of the small spectral gap discussed in Sec.~\ref{sec:spectral}.

\section{Discretizing the Fokker--Planck Operator} \label{app:discretization}

Our goal is to discretize the FP operator $\mathcal{L}^\dagger$ for numerical eigenvalue and eigenvector analysis. We follow an upwind scheme \cite{leveque2002finite} proposed in \cite{lin2022koopman}, which turns the FPE into a continuous-time Markov chain on the discretized domain, structurally conserving the probability mass. Specifically, we truncate the state space in $x\in \left[-L,L\right]$ and uniformly discretize the interval into $N=2000$ grid points, $-L=x_1 < x_2 < \ldots x_N = L$, $x_{i+1}-x_i = \Delta :=2L/(N-1)$, $i=1\ldots N-1$. In addition, we will impose reflective boundary conditions at $x=L$ and $x=-L$ to ensure conservation of probability. 

We first define the ``right'' and ``left'' transition rates at grid $j$ as
\begin{align}
    & l_{j} = \frac{1}{\Delta^2 } - \frac{1}{2 \Delta} v_j, \quad  r_{j} = \frac{1}{\Delta^2} + \frac{1}{2 \Delta} v_j,
\end{align}
where $v_j:= \left(\text{d}/\text dx\right) \log \nu(x_j)$ is the drift in Eq.~\eqref{eq:fSDE} evaluated at grid $j$. 
Then, the Fokker--Planck operator $\mathcal{L}^\dagger$ defined in Sec.~\ref{sec:spectral} can be discretized as an $N\times N$ matrix $L^\dagger$ whose elements are
\begin{align} \label{eq:discrete_L}
    &[L^\dagger]_{ij} = \begin{cases}
l_j & \text{if } 2 \leq j \leq N-1 \ \text{and} \ i=j-1, \\
-(l_j+r_j) & \text{if } 2 \leq j \leq N-1 \ \text{and} \ i=j,  \\
r_j & \text{if } 2 \leq j \leq N-1 \ \text{and} \ i=j+1, 
\end{cases} 
\end{align}
with reflective boundary conditions:
\begin{align}\label{eq:discrete_L_bc}
    &[L^\dagger]_{ij} = \begin{cases}
-r_j & \text{if } (i,j)=(1,1),  \\
r_j & \text{if } (i,j)=(2,1),  \\
l_j & \text{if } (i,j)=(N-1,N), \\
-l_j & \text{if } (i,j)=(N, N). 
\end{cases}
\end{align}

\section{Experiment Specification}
\label{app:experiment-specification}

\begin{figure*}[!t]
    \centering
    \includegraphics[width=\linewidth]{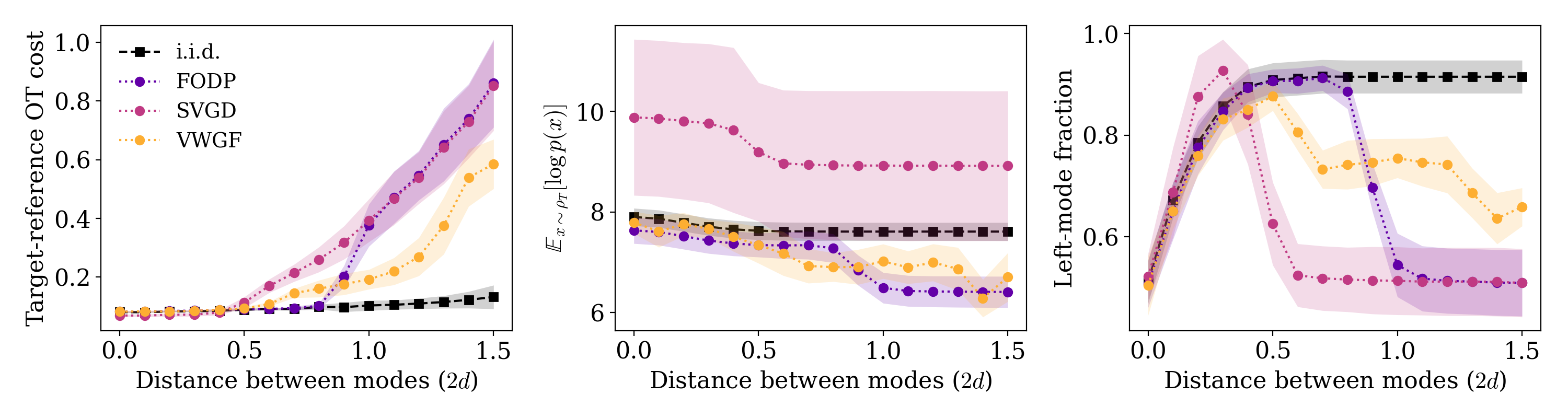}
    \caption{Performance comparison of different WGF-based samplers and i.i.d.~samples of the bimodal Gaussian mixture in $D=10$.}\label{fig:sampleQualities10D}
\end{figure*}

This appendix specifies the controlled two-mode Gaussian-mixture experiments
used to generate the one-dimensional and ten-dimensional results in
Figures~\ref{fig:sampleQualities} and
\ref{fig:sampleQualities10D}. The comparison includes Stein Variational
Gradient Descent (SVGD), forward SDE sampling via overdamped Langevin dynamics
(FODP), Variational Wasserstein Gradient Flow (VWGF), and exact i.i.d.~sampling
from the target mixture.

\subsection{Target Distribution}

For each dimension \(D \in \{1,10\}\), the target distribution is the
two-component isotropic Gaussian mixture
\[
    p(x)
    =
    \alpha\,\mathcal{N}(x;\mu_L,\sigma^2 I_D)
    +
    (1-\alpha)\,\mathcal{N}(x;\mu_R,\sigma^2 I_D),
    \qquad x \in \mathbb{R}^D.
\]
The component means are separated only along the first coordinate:
\[
    \mu_L=-d e_1,
    \qquad
    \mu_R=d e_1,
    \qquad
    s=2d,
\]
where \(s\) is the plotted full mode separation and
\(e_1=(1,0,\ldots,0)^\top\).

The unnormalized component weights are \(w_L=10\) and \(w_R=1\), so that
\[
    \alpha = \frac{w_L}{w_L+w_R}=\frac{10}{11},
    \qquad
    1-\alpha=\frac{1}{11}.
\]
The component variance is fixed at $\sigma^2=0.012$.

\begin{center}
\begin{tabular}{lll}
\toprule
Quantity & Symbol / Name & Value \\
\midrule
Dimensions & \(D\) & \(1\), \(10\) \\
Left/right unnormalized weights & \(w_L,w_R\) & \(10,1\) \\
Left/right mixture masses & \(\alpha,1-\alpha\) & \(10/11,1/11\) \\
Component variance & \(\sigma^2\) & \(0.012\) \\
Component covariance & -- & \(\sigma^2 I_D\) \\
Full separation grid & \(s\) & [0.0,0.1,0.2,...,1.5]\\
Half separation & \(d\) & \(s/2\) \\
\bottomrule
\end{tabular}
\end{center}

All score-based methods use the analytic score of this Gaussian mixture. The
same target definition is used for both \(D=1\) and \(D=10\).

\subsection{Initialization}

For each target setting and random seed, particle-based methods are initialized
from a standard normal distribution:
\[
    x_0^{(i)} \sim \mathcal{N}(0,I_D),
    \qquad i=1,\ldots,N.
\]
Thus, the initialization is independent of the target mode locations and does
not initialize particles in either mixture component.

\subsection{Methods}

{\bf SVGD:} The SVGD experiments use \(N=100\) particles and \(3000\) iterations with step
size \(\epsilon=10^{-3}\). The kernel is the RBF kernel
\[
    k(x,y)=\exp\left(-\frac{\|x-y\|^2}{h^2}\right).
\]
At each target setting, SVGD is run for \(20\) independent replicates. The RBF
bandwidth is varied across replicates by sampling
\[
    \log h \sim \mathrm{Uniform}(\log 0.05,\log 1.0).
\]
AdaGrad-style scaling is enabled with historical gradient decay \(0.9\), fudge
factor \(10^{-6}\), finite-value checks, and maximum particle magnitude
\(10^6\).

{\bf FODP:} The FODP baseline uses Eq.~\eqref{eq:fSDE},
\[
    dX_t=\nabla \log p(X_t)\,\text{d}t+\sqrt{2}\,\text{d} W_t,
\]
discretized by Euler--Maruyama with step size \(h_{\mathrm{SDE}}=1\times10^{-3}\).
The final time is \(T=10\), giving \(10000\) discretization steps and per-step
noise scale \(\sqrt{2h_{\mathrm{SDE}}}=\sqrt{0.002}\). FODP is run for \(20\)
independent replicates at each target setting, using \(N=100\) particles.

{\bf IID:} The exact i.i.d.~baseline draws independent samples directly from the target
mixture. Unless otherwise stated, the i.i.d. baseline uses \(N=100\) samples per
replicate and \(20\) independent replicates.

{\bf VWGF:} The VWGF experiments were generated using the implementation and
hyperparameters summarized in Table~\ref{tab:vwgf-spec}. For each target
setting, VWGF was trained once with 512000 particles. After training, \(20\) terminal 100-particle clouds
were generated via subsampling. These terminal clouds were then evaluated using
the same metrics as the other methods.

\begin{table}[ht]
\centering
\begin{tabular}{lll}
\toprule
VWGF quantity & Symbol / Name & Value \\
\midrule
Training particle count & \(N_{\mathrm{train}}\) & 512000 \\
Evaluation particle count & \(N_{\mathrm{eval}}\) & \(100\) \\
Batch size & -- & 512 \\
Number of training epochs & -- & 40 \\
Learning rate & -- & 0.001 \\
Network architecture --- layers & -- & 2 (D=1), 3 (D=10) \\
Network architecture --- hidden dimension & -- & 8 (D=1), 32 (D=10) \\
Training replicates per target setting & -- & \(1\) \\
Evaluation replicates per trained VWGF model & -- & \(20\) \\
Evaluation initialization & -- & \(x_0^{(i)}\sim\mathcal{N}(0,I_D)\) \\
\bottomrule
\end{tabular}
\caption{VWGF settings used for the appendix comparison.}
\label{tab:vwgf-spec}
\end{table}

\subsection{Metrics}

All methods are evaluated using terminal samples or terminal particles. For a
sample cloud \(\{x_i\}_{i=1}^N\), the mean log probability is
\[
    \frac{1}{N}\sum_{i=1}^{N}\log p(x_i).
\]
This metric measures whether particles lie in high-density regions, but it does
not by itself determine whether the mixture weights are represented correctly.

For \(s>0\), the empirical left-mode fraction is
\[
    \widehat{\alpha}
    =
    \frac{1}{N}
    \sum_{i=1}^{N}
    \mathbf{1}\{x_{i,1}<0\}.
\]
The target value is \(\alpha=10/11\). At \(s=0\), the two component means
coincide, so sign-based mode assignment is not geometrically meaningful.

The optimal transport metric compares the empirical sample measure
\[
    \widehat{p}
    =
    \frac{1}{N}
    \sum_{i=1}^{N}\delta_{x_i}
\]
to an independent exact reference sample from the target,
\[
    \widehat{p}_{\mathrm{ref}}
    =
    \frac{1}{N_{\mathrm{ref}}}
    \sum_{j=1}^{N_{\mathrm{ref}}}\delta_{y_j},
    \qquad
    y_j\sim p.
\]
The reported metric is the squared Wasserstein-2 distance between
\(\widehat{p}\) and \(\widehat{p}_{\mathrm{ref}}\) using squared Euclidean cost
\(c(x,y)=\|x-y\|_2^2\).

\section{Effect of Extended Simulation Time}

We further investigate the effect of increasing the terminal simulation time $T$ for the forward-only diffusion process (FODP) in one-dimensional settings. In particular, we repeat the $D=1$ experiments with progressively larger values of $T$ while keeping all other parameters fixed.

As expected from the spectral and MFPT analyses, increasing $T$ improves the ability of the sampler to explore separated modes. This improvement can be attributed to the intrinsic stochasticity of the diffusion process: inter-mode transport occurs through rare transition events, which correspond to the slowest decaying eigenmode of the Fokker--Planck operator. In the metastable regime, these transitions are governed by transition-path and large-deviation theories, and their occurrence probability increases with simulation time. Consequently, extending $T$ allows a larger fraction of sample trajectories to overcome energy barriers and contribute to correct mode occupancy.

Empirically, we observe that for fixed mode separation, the quality of the sampled distribution improves gradually as $T$ increases. The improvement is consistent with the mixing time estimation in Fig.~\ref{fig:mixingTime}. In particular, the estimated mode weights become closer to the ground truth and the Wasserstein distance decreases. However, this improvement is extremely slow: the required simulation time to achieve well-mixed samples grows rapidly with the separation between modes. For moderately separated modes, achieving accurate mixing may require values of $T$ several orders of magnitude larger than those considered practical in typical applications.

This behavior is consistent with the theoretical prediction that the mixing time scales exponentially with the barrier height (or equivalently, mode separation). Therefore, while increasing $T$ can in principle recover correct global statistics for FODP, it does not resolve the fundamental inefficiency of gradient-driven dynamics in multimodal settings. Instead, it highlights that the limitation is not a failure of convergence, but a consequence of exponentially slow mixing. 

We remark that deterministic WGF-based samplers, such as SVGD and VWGF, do not benefit from increasing the terminal horizon $T$ beyond stabilization. As discussed in Sec.~\ref{sec:discussion}, their purely deterministic relaxation dynamics lead to stabilization within local modes, and extending $T$ does not introduce additional inter-mode transitions or improve global mixing.


\end{document}